# Human-Centric Aware UAV Trajectory Planning in Search and Rescue Missions Employing Multi-Objective Reinforcement Learning with AHP and Similarity-Based Experience Replay*

Mahya Ramezani and Jose Luis Sanchez-Lopez

*Abstract*— The integration of Unmanned Aerial Vehicles (UAVs) into Search and Rescue (SAR) missions presents a promising avenue for enhancing operational efficiency and effectiveness. However, the success of these missions is not solely dependent on the technical capabilities of the drones but also on their acceptance and interaction with humans on the ground. This paper explores the effect of human-centric factor in UAV trajectory planning for SAR missions. We introduce a novel approach based on the reinforcement learning augmented with Analytic Hierarchy Process and novel similarity-based experience replay to optimize UAV trajectories, balancing operational objectives with human comfort and safety considerations. Additionally, through a comprehensive survey, we investigate the impact of gender cues and anthropomorphism in UAV design on public acceptance and trust, revealing significant implications for drone interaction strategies in SAR. Our contributions include (1) a reinforcement learning framework for UAV trajectory planning that dynamically integrates multi-objective considerations, (2) an analysis of human perceptions towards gendered and anthropomorphized drones in SAR contexts, and (3) the application of similarity-based experience replay for enhanced learning efficiency in complex SAR scenarios. The findings offer valuable insights into designing UAV systems that are not only technically proficient but also aligned with human-centric values.

## I. INTRODUCTION

Unmanned Aerial Vehicles (UAVs), commonly known as drones, have emerged as a forefront technology in autonomous systems, demonstrating significant advancements across various industries [1]. Despite the technical progress, the integration of UAVs into societal operations necessitates a deeper understanding of human-centric factors that influence this interaction, especially within the domain of Human-Drone Interaction (HDI) [2]. In the field of Human-Robot Interaction (HRI) [3], valuable insights have been provided into how the physical and behavioral attributes of robots affect human interactions [4]. To enhance robots' acceptance in HRI, incorporating anthropomorphism and human social traits improves familiarity and acceptance [5]. Anthropomorphism, or the addition of human-like features to robots, enriches their design with social cues and facilitates interaction, enhance connection and acceptance [4, 6]. Aligning a robot's appearance and capabilities with user expectations further boosts acceptance [7].

One crucial aspect of anthropomorphism is gender, which significantly influences acceptance and efficiency in HRI. This factor not only impacts the roles assigned to robots but also affects human acceptance and trust in specific roles [3].

Anthropomorphism significantly impacts public perception and interaction with drones [8, 9]. Drones featuring social traits, such as a voice or a face, decrease perceived distance for human interaction, thus improving acceptance. This is especially vital for missions like Search and Rescue (SAR), which prioritize human-drone interaction.

Integrating human factors and emotional cues in drones significantly enhances the efficacy of SAR missions, fostering improved engagement and interactions [10]. Recent research emphasizes the deployment of behavior-based Artificial intelligence in UAVs for SAR, highlighting their autonomous capabilities for efficient disaster response [11]. Advanced maneuverability, reliability, and sensing are key to their success in navigating complex terrains and detecting survivors. Another study showcases UAVs' strategic advantage in SAR operations, underlining the importance of intuitive human-drone interactions and the potential of autonomous drones to elevate operational efficiency and success in challenging environments [9, 12].

Despite existing research on HDI, there remains a noticeable gap regarding the influence of anthropomorphism, particularly gender cues, on UAV design and trajectory planning within HDI contexts. This study aims to bridge this gap by exploring if principles noted in HRI are equally pertinent to HDI, especially concerning the gender perception of drones as non-humanoid robots. Factors such as size, design, and color play pivotal roles in anthropomorphizing drone designs and influencing gender perceptions [13].

Our research investigates the correlation between drones' physical and behavioral attributes and the success of SAR missions. We analyze how UAV design impacts efficiency in SAR operations, as well as human preferences in mission trajectory planning for success.

This study underscores the importance of gender cues in drone design, examining their impact on mission efficiency, role suitability, and human preferences. It contributes significantly to integrating gender considerations into UAV design and operation within HDI frameworks. The analysis delves into the practical implications of design choices in SAR operations. the study evaluates how anthropomorphic and gender cues in drone design may enhance effectiveness and acceptance, aiming to improve mission success and comfort for all participants. Another crucial aspect of SAR success involves considering human-centric factors and implementing autonomous trajectory planning while prioritizing human-centered approaches. To achieve this, we employ a Deep

*The authors are with the Automation and Robotics Research Group, Interdisciplinary Centre for Security, Reliability and Trust (SnT), University of Luxembourg (UL).
M. Ramezani (corresponding author; e-mail: mahya.ramezani@uni.lu)
J. L. Sanchez-Lopez (e-mail: joseluis.sanchezlopez@uni.lu)

Reinforcement Learning (DRL) algorithm tailored for SAR missions [1], which autonomously and adaptively enhances efficiency across various objectives during SAR operations. Our research progresses through a meticulously designed survey, aiming to unravel the general public's perception of gender cues in drones.

A pivotal contribution of this paper is the development of a trajectory planning framework for UAVs in SAR missions that integrates human-centric factors. This approach aims to synchronize operational goals with human-centric considerations, enabling more intuitive, adaptive, and effective UAV deployment in critical rescue operations.

- The study investigates the strategic utilization of gender perceptions to optimize UAV design through survey.
- We utilize a DRL-based algorithm for trajectory planning, prioritizing human-centric factors, energy efficiency, and time efficiency. We incorporate the Analytic Hierarchy Process (AHP) for dynamic reward allocation and introduce similarity-based experience replay to address sample efficiency and non-stationary issues.

## II. SURVEY: ASSESSING THE IMPACT OF GENDER CUES IN DRONE DESIGN ON ROBOT ACCEPTANCE AND ROLE ASSIGNMENT IN SAR MISSIONS

This section examines how gender cues and anthropomorphism in UAV design affect human trust and acceptance in various operational contexts, particularly in SAR missions. The efficacy of SAR missions relies heavily on human-drone interaction. The study aims to analyze how different levels of anthropomorphic design, especially gender cues, impact drone operational effectiveness in SAR tasks. It investigates the effects of design attributes such as gender-specific vocalizations, changes in movement speed, and gendered physical appearances on human perceptions of safety, comfort, and interaction politeness, which in turn may influence preferred interaction distances. The section aims to:

- Determine public perceptions of drone gender based on assigned gendered adjectives and traditional roles.
- Evaluate preferences for drones with specific genders in various roles, particularly SAR tasks.
- Implement gender-specific design elements, voices, and adjust drone speed and altitude to enhance interaction safety.

To accomplish these goals, the study altered the design of various drones through visual and behavioral characteristics, including size, color, and design elements typically associated with masculinity or femininity. This methodological approach is underpinned by literature [14, 15] that delves into gendered adjectives and their application in design.

### A. Survey Instrument

In this survey, advanced artificial intelligence-based generative models, notably DALL·E-2 [16], were utilized to progressively refine the design of UAVs. These models were tasked with generating varied versions of a commercial drone, as illustrated in Fig. 1, encompassing designs that were more distinctly feminine, masculine, and an animal-shaped (canine) variant, alongside the original configuration.

For each design variant, a 5-second video was created in which the drone introduced itself against various backgrounds. A consistent dialogue was used across all versions: "I am a drone. I can assist you in various applications and possess numerous capabilities." Voiceovers for each variant were distinct: the masculine version featured a male voice, the feminine version a female voice, the canine-shaped design communicated through barks with accompanying subtitles, and the original design used a gender-neutral voice. In addition, the original design was presented also, without voice narration, the latter including subtitles to convey its capabilities, which is "This is a DJI drone, it can assist you in various application and possess numerous capabilities." Participants in the study were randomly assigned to one of the five categories. Subsequently, they were asked to complete a survey, which on average took less than 5 minutes.

Furthermore, the survey included video and images snippet for each drone variant to showcase the UAV's operational capabilities in post-disaster scenarios, featuring different speeds. It should be noted that the survey was conducted online.

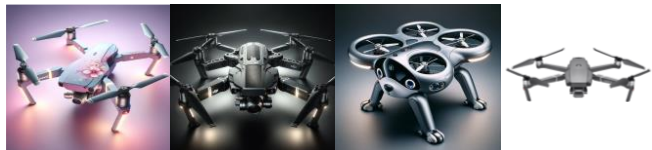

Figure 1. The different variant of drone's design from left drone A, drone B, drone C, and drone D.

### B. Pilot Study

The pilot study aims to validate survey parameters and assess the main hypothesis. Before the main investigation, an initial pilot study was conducted in person and involved participants within the age range of 25 to 55 years. Participants were asked to express the gender they attributed to a drone, scale from 1 to 5, where 1 signified a stronger feminine perception.

To validate the appropriateness of our study's chosen attributes and occupations, participants rated each adjective based on its typical association with males or females. The results, with low standard deviations, indicated strong participant agreement. Mean ratings generally aligned with societal norms and accepted gender stereotypes. Also, the participants were asked to rate the perceived gender of various voice samples, and drone designs for efficiency of the survey instrument. Results validated the parameters.

Many participants hesitated to respond directly to the question, often resorting to humor or dismissiveness. To address this, we adopted an indirect approach inspired by [17] to explore how individuals attribute gender to drones.

### C. Survey Detail

In the survey, participants first evaluated drones based on a set of attributes representing masculinity and femininity, as outlined in [18]. They then assessed the drones' suitability for ten specific occupations, equally divided between traditionally male and female roles [19]. Additionally, participants were asked about their comfort and perceived safety in close proximity to each drone design, including their willingness to allow drones to land on their hands. To evaluate perceptions of drones' efficacy in SAR missions, participants ranked the likelihood of successful mission completion in various scenarios. They also provided feedback on drones' speeds (5, 20, 50 km/h) and their impact on human proximity and comfort levels, as well as hypothetical reactions of survivors to encountering drones' post-disaster. Participants also shared their willingness to seek assistance from drones in disaster aftermath situations and potential aggressive responses towards drones in case of operational errors. It should be noted that all the scale is based on the Likert scale Likert scale [4] from 1 (lowest)-10 (highest). Finally, participants rated the perceived gender of the drones on a scale of 1 to 10, where 1 represented a feminine perception, 10 a masculine perception, and 5 gender-neutral. Participants' personal details, including age and gender, were also collected.

*D. Results and discussion*

The survey included 150 participants, 56% male and 44% female, aged 18-60, showed diversity with Asian (20.6%), Middle Eastern (39.4%), European (28.6%), African (5.4%), and Latino (6%) backgrounds. English fluency was confirmed for the English-language survey.

The drones were categorized into five distinct design groups, and each survey was conducted separately for each category to prevent design bias from influencing participants' responses. These categories included Drone A, featuring a feminine design with a female voice; Drone B, characterized by a masculine design with a male voice; Drone C, designed to resemble a canine shape with barking sounds and subtitles; Drone D1, presenting an original design with a neutral voice; and Drone D2, showcasing the same original design without voice (see Fig. 1). The results indicate that participants perceived Drone A as more feminine (Mean Score (MS) = 3.5, Standard Deviation (SD) = 2.58), Drone B as more masculine (MS = 8.33, SD = 1.83), and Drone C as neutral (MS = 5.58, SD = 1.25). Similarly, Drone D1 was also perceived as neutral (MS = 5.88, SD = 2.05). In contrast, Drone D2 was perceived as more masculine than neutral (MS = 6.5, SD = 1.05).

These findings demonstrate that the addition of gender cues and anthropomorphism in design can significantly influence the perception of non-humanoid robots among participants, leading them to assign gender attributes to these robots. Furthermore, the inclusion of a neutral voice in drones, as opposed to those without voice, appears to reduce their perceived masculinity and promote a more neutral perception among participants. The results also reveal that the canine design was perceived as more neutral by the participants, suggesting that less anthropomorphism leads to a more neutral gender perception.

Fig. 2 displays the average scores for various robot attributes, encompassing five attributes each that are traditionally associated with male and female characteristics. The findings indicate that Drone A obtained a mean score of 5.738 for Traditionally Female (TF) attributes and 4.45 for Traditionally Male (TM) attributes. Drone B received a mean score of 4.56 for TF attributes and 6.7 for TM attributes, while Drone C received mean scores of 5.434 for TF and 5.1 for TM attributes. Similarly, Drone D1 received mean scores of 4.36 for TF attributes and 5.153 for TM attributes, and Drone D2 had a mean score of 4.85 for TF attributes and 5.55 for TM attributes. The results suggest that Drone B, characterized by male attributes, excels in masculine attributes, whereas Drone A, embodying feminine attributes, demonstrates higher mean scores for female attributes. Additionally, the comparison among other drones reveals similar averages, indicating a potential relationship between gender cues in design and attribute assignment. Notably, in non-humanoid robots like drones, gender cues in design could influence attribute perception.

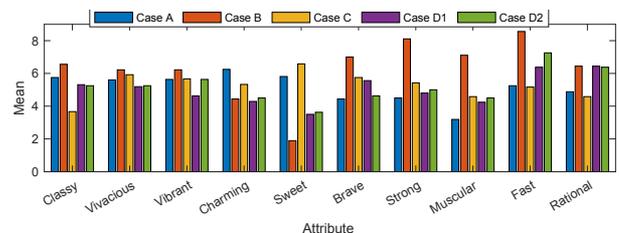

Figure 2. The TM and TF attributes. The first five attributes are TF, while the others are TM.

Fig. 3 shows trust levels in traditionally gendered occupations by drone, with four male and the rest female roles, validated in a pilot study. Drone A averages a 6.03 trust level, B scores 6.84, C 6.46, D1 6.12, and D2 6.61. The data suggest masculine design cues in drones correlate with higher trust in task efficiency, with Drone B leading in perceived efficiency across all roles, especially those considered masculine, and D1 following closely. Drone C is noted for excelling in specialized tasks due to its design, indicating specific designs can enhance trust in certain contexts.

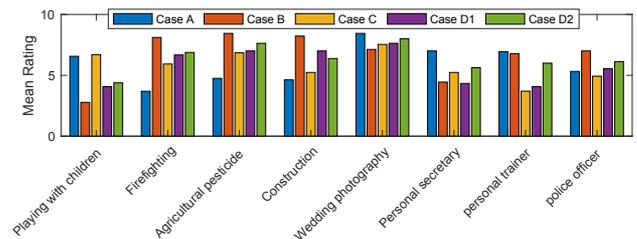

Figure 3. Trust for each drone design in TM and TF occupations.

The study shows different reactions to drone use in gender-typical roles and SAR tasks. Drone A was preferred for TF roles, scoring 7.125, over SAR (5.71875) and TM roles (5.0625). Drone B favored TM and SAR uses, with scores of 7.7111 and 8.5556, against TF's 5.2917, indicating a preference for more demanding or urgent uses. SAR was well-received across drones, especially Drone B (8.5556) and Drone D2 (7.5625). This variation highlights drones' versatile uses, shaped by societal standards and context, and shows how gender-related design cues affect trust in TF and TM occupations.

As shown in Fig. 4, participants exhibited the highest comfort level interacting with the Drone A, with an MS of 7.125, and were most willing to allow the drone to land on their hand, indicating a significant level of trust in its safety and functionality. Drone B showed a slightly higher inclination to consider purchasing (4.875) and paying more, reflecting trust in its efficiency, though participants were more ambivalent about interaction comfort. Drone B also displayed the highest pricing agreement mean (6.6667), suggesting recognition of its value despite safety concerns, indicated by lower safety feeling ratings (4.8889). Drones C, D1, and D2 presented mixed but generally moderate views on interaction comfort and likelihood of drone purchase. Participants in Drone D1 were notably less inclined to consider purchasing the drone (3.9333) and expressed the highest likelihood of feeling angry if it malfunctioned (6.0), indicating frustration. Safety perceptions and willingness to allow the drone to land on one's hand varied, reflecting ongoing reservations about drone proximity and interaction safety. Data suggests that the D1 neutral configuration is most likely to induce anger and provoke negative behavior, highlighting the influence of gender cues in design and people's willingness to accept them.

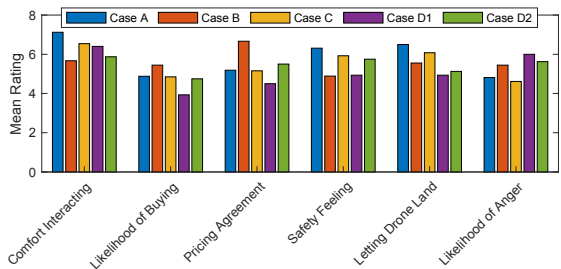

Figure 4. Human acceptance based on different drone designs.

As illustrated in Fig. 5, the assessment of people's willingness to use drones for SAR missions reveals intriguing insights into the perceived gender cues of the drones and their impact on participant responses. Case B, associated with more traditionally masculine gender cues, stands out as the scenario where participants expressed the highest likelihood of using the drone as a teammate in SAR competitions, with a mean rating of 8.6667. This contrasts with perceptions towards drones in Drones A and D1, where the MS are 6.375 and 6.875, respectively. Notably, while Drone B is considered to have more masculine cues, it still received a higher willingness rating to be used as a pilot (7.45) compared to other cases. Furthermore, when examining the likelihood of purchasing the drone for disaster assistance, Drones B, C, and D1 are among the highest, with MS of 6.7778, 6.7692, and 6.0667, respectively, showcasing a close range of responses suggesting that a more masculine perception could positively influence the decision to purchase drones for use in SAR missions. Specifically, Drone B not only rates highest in terms of being used as a teammate (8.6667) but also shows a significant openness to being piloted in real SAR missions (7.4444) and a robust willingness to purchase for disaster scenarios (6.7778).

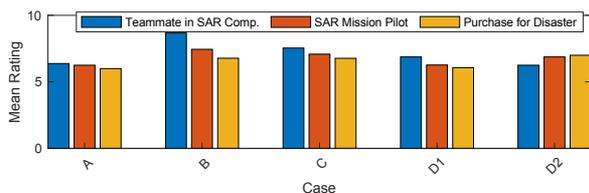

Figure 5. The willingness to engage with different designs in SAR missions.

These findings underscore a relationship between the perceived gender cues of drones and their acceptance for critical roles in SAR missions. The data suggests that drones associated with more masculine traits, as reflected in Case B, tend to inspire greater confidence in their capabilities for teamwork and piloting in SAR contexts, as well as a higher propensity among participants to consider purchasing them for disaster assistance. This indicates that perceptions of masculinity in drone design or presentation could have substantial effect on their perceived utility and desirability in life-saving operations.

The analysis presented in Fig. 6 elucidates the relationship between drone speed and the likelihood of individuals running away when the drone approaches near participants at varying speeds. The results underscore a clear trend: as the drone's speed increases, so does the likelihood of individuals choosing to run, highlighting the significant impact of drone velocity on human behavior during critical missions. This observation is crucial, emphasizing that even when drone assistance is vital, the approach speed can influence the receptiveness and comfort levels of those being aided.

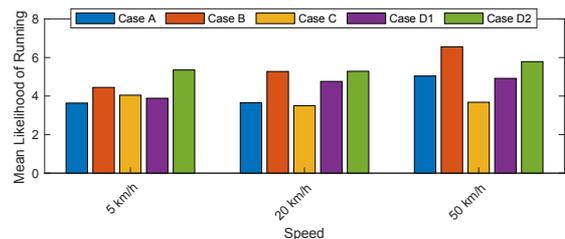

Figure 6. Likelihood of humans fleeing from different drone designs approaching at varying speeds.

Notably, Drone C, characterized by a drone design inspired by animal shapes and animation, demonstrates a lower likelihood of individuals running from it. This design choice seems to evoke a sense of sympathy and friendliness towards the drone, making it appear more approachable and less intimidating to humans. Similarly, Drone A, which incorporates feminine gender cues in its design, also shows a trend where individuals are more tolerant and possibly more receptive to its presence. This tolerance suggests that design cues significantly influence human interaction with drones, with more sympathetic or relatable designs leading to better acceptance and less fear during interactions. Additionally, D1, which is equipped with a neutral voice, demonstrates that, unlike D2, there may be a lower likelihood of participants fleeing from it.

Results show that, the Drone A were found to be more comforting, with participants expressing a higher tolerance for closer proximity at lower speeds. This suggests that designs perceived as more nurturing or gentle may enhance the willingness of individuals to engage closely with drones, particularly in non-threatening scenarios. The drone B cues tended to be associated with a higher likelihood of individuals running at increased speeds, reflecting perhaps a subconscious association of masculine features with strength or aggression. Despite this, the masculine design did not deter close interactions at lower speeds, indicating that any perceived intimidation might be mitigated by controlled operational parameters. The dog-shaped drone was notably effective in reducing the likelihood of running, regardless of speed, underscoring the positive impact of approachable, non-humanoid shapes on human-drone interactions. This design's sympathetic and familiar appeal likely contributes to its acceptance across various operational speeds. The presence of a neutral voice in D1 did not significantly alter proximity preferences compared to D2, where no voice was presented. However, auditory feedback could influence comfort levels in nuanced ways not fully captured by proximity preferences alone, potentially affecting perceptions of the drone's approachability and the intuitiveness of human-drone interactions. The lower likelihood of running from drones with animal-inspired designs or feminine cues highlights the importance of drone appearance in human-drone interaction. These design elements can make drones appear less threatening and more relatable, potentially increasing their acceptance in sensitive scenarios.

Across all drone design, there was a clear trend that increased drone speed resulted in a preference for greater distances, highlighting safety concerns. This effect was most pronounced at the highest speed (50 km/h). The increased likelihood of running away from faster-moving drones underscores a fundamental tension between operational efficiency and human comfort. While faster drones can cover more ground quickly, their speed may counterintuitively decrease their effectiveness in missions requiring close interaction with people. Therefore, balancing these aspects is crucial for mission success. Additionally, research indicates that female participants tend to assign gender to drones more than males, who often perceive them as neutral.

Our study reveals how design, anthropomorphism, and gender cues critically influence human-drone interactions, especially in SAR missions. It highlights the importance of integrating human-centric design principles to enhance user perception and trust. Anthropomorphic features and gender cues in drones significantly affect user expectations of their capabilities, aligning with theories of anthropomorphism, social role and emotional design which together suggest that relatable and emotionally positive design enhances user acceptance and cooperation. Furthermore, gender schema theory shows how gender perceptions shape interactions, emphasizing the need for diverse design considerations [20].

The study's findings underscore the necessity for designs that not only foster trust and comfort but also meet varied user expectations, highlighting the role of design in successful human-drone interactions for SAR missions. Addressing human preferences in drone speed and trajectory planning is crucial for safety and effectiveness, suggesting a move towards multi-objective DRL strategies for better human-drone synergy. The interplay between design aesthetics and perceived drone functionality demonstrates a preference for designs that balance efficiency with user comfort, indicating a broader impact on trust and acceptance across different tasks and contexts.

### III. AWARE TRAJECTORY PLANNING FOR SEARCH AND RESCUE MISSION

This section advances autonomous UAV trajectory planning for SAR missions, integrating multi-objective considerations to harmonize operational efficiency with human-centric objectives. By optimizing path planning for effective food delivery in post-disaster scenarios. The UAV is outfitted with LIDAR technology for navigating around obstacles and utilizes GPS for precise positioning.

In this section, we address a multi-objective optimization challenge for UAV trajectory planning in SAR missions, focusing on balancing time efficiency, energy efficiency, obstacle avoidance, and adherence to human preferences include safe proximity and safe speed relative to survivors. Our objectives include minimizing mission duration and energy consumption, ensuring effective navigation around obstacles detected by radar, and maintaining survivor comfort by adjusting the UAV's proximity and speed.

Formally, the optimization task is defined by the multi-objective function $F(S, A) = [f_1, f_2, f_3, f_4]$, where each $f_i$ corresponds to one of objectives. We propose a reinforcement learning strategy that scalarizes $F$ into a singular objective through dynamic weighting, allowing for real-time adjustments based on environmental and survivor conditions. This method ensures a harmonized approach to UAV trajectory planning, optimizing for operational efficiency and survivor safety concurrently. The mathematical model of the UAV, detailed in [21], is omitted here for conciseness.

*A. problem modeling*

The operational environment is modeled as a Markov Decision Process, characterized by continuous state and action spaces. An MDP is defined by a tuple $(S, A, P, R, \gamma)$, where '$S$' denotes the state space, '$A$' represents the action space, '$P: S \times A \times S \to [0,1]$' specifies the transition probability function $P(s'|s, a)$, indicating the likelihood of transitioning from the current state '$s$' to a new state '$s'$' upon executing action '$a$'; '$\gamma$' is a discount factor within the range $(0, 1)$, and '$R: S \times A$' defines the reward function.

State spaces encompasses a set of variables describing the UAV's environment and operational status. We model UAV navigation within a two-dimensional plane using a simplified kinematic model, assuming constant altitude and linear velocity. The state of the UAV is characterized by its position $(x, y)$, velocities $(v_x, v_y)$, orientation angle $\theta$, and battery level $(B)$. obstacle sensing is achieved through laser distance sensors with a comprehensive scan capability. Additionally, we incorporate the survivor's velocity $(v_s)$ and position $(x_s, y_s)$ Mathematically, the UAV state set is defined as:

$$S = \{(x, y), (v_x, v_y), \theta, B, vs, (x_s, y_s), d_0, \dots, d_n\} \quad (1)$$

The UAV's action set is included to two primary inputs: speed and yaw angle, encapsulated within the control vector $u = [u_1, u_2]$. Here, $u_1$ specifies the ratio of current speed to maximum speed, adopting values within the interval $[-1,1]$, and $u_2$ represents the steering command that dictates the desired yaw angle, also ranging from $[-1,1]$.

Reward function is composed of distinct components, each tailored to a specific aspect of the mission:

– **Time Efficiency ($R_t$):** Rewards faster mission completion with $R_t = -\Delta t$, where $\Delta t$ is the time step.
– **Energy Efficiency ($R_e$):** Penalizes energy depletion proportionally to $R_e = \frac{B}{B_c - B}$, promoting efficient energy use where $B_c$ is the battery capacity.
– **Obstacle Avoidance ($R_o$):** Encourages distance from obstacles $R_o = exp(-d_n)$ for $d_n < 20$, where $d_n$ is the distance to the nearest obstacle with penalties for collisions.
– **Human Preferences ($R_h$):** Penalizes unsafe proximity and speeds, particularly $R_h = - \times |v_s|$ when inducing survivor distress, emphasizing gentle interaction. Unsafe proximity is defined as being within 2 m, while safe proximity is 3.5 m.

The total reward $R(s_t, a_t)$ at each timestep $t$ is

$$R(s_t, a_t) = w_t \cdot R_t + w_e \cdot R_e + w_o \cdot R_o + w_h \cdot R_h, \quad (2)$$

where $w_t, w_e, w_o,$ and $w_h$ are dynamic weights reflecting the relative importance of each time, energy, obstacle avoidance and human preference objectives based on the UAV's current state. The maximum speed of the UAV is 22 m/s, and it flies at a constant altitude of 10 m.

*B. Proposed Method*

We employ The Twin Delayed Deep Deterministic policy gradient (TD3) algorithm [22], augmented with the Analytic Hierarchy Process (AHP), to dynamically prioritize objectives within our UAV trajectory planning framework. The reward function for each state is dynamically shaped by these AHP-derived weights, allowing for an adjustment of rewards that aligns with the mission's evolving priorities. To overcome non-stationarity in learning for the dynamic reward and improve learning efficiency, we implement a similarity-based experience replay mechanism. This approach ensures the strategic replay of experiences that are contextually relevant to the current state, significantly improving the learning process and efficiency.

*1) Twin Delayed Deep Deterministic policy gradient*

The TD3 algorithm is an advanced reinforcement learning method designed to address the inherent challenges associated with the Deep Deterministic Policy Gradient (DDPG) algorithm. Recognizing the susceptibility of DDPG [1] to overestimation bias and its subsequent impact on stability and performance, TD3 introduces three key enhancements: the use of twin critics, delayed policy updates, and target policy smoothing. These modifications collectively contribute to TD3's superior robustness and efficiency, making it particularly suitable for applications requiring precise and reliable control in complex environments, such as UAV trajectory planning for SAR missions. This approach, combined with infrequent actor updates and added noise to target actions, ensures more accurate value estimations and a more stable learning process. [22].

*2) Dynamic Reward weighting using AHP*

To adeptly balance competing objectives, we apply AHP to dynamically assign weights to each component of the reward function based on states. AHP, a structured technique for organizing and analyzing complex decisions, is particularly suited for this task, allowing for the integration of expert judgment and empirical data in determining the relative importance of each objective. This process is contingent upon the UAV's current state, categorized by its proximity to obstacles and survivors, and specific mission phases.

In our method, objectives are closely tied to the distance to the survivor and the proximity to obstacles. These factors significantly influence the objective of our problem. Therefore, we categorize states based on their proximity to obstacles and survivors. This categorization is crucial for setting the context of the learning scenario and guiding how rewards will be distributed in the subsequent stages.

The categorization is based on the Lidar's threshold range.
- Far from survivor and obstacle ($l_1$)
- Near from the obstacle and survivor ($l_2$)
- Near from survivor, far from obstacle ($l_3$)
- Near from obstacle, far from survivor ($l_4$)

For each categorized state, the AHP is employed to determine the most appropriate weight for each state successes in its objectives to reward structure. This results in a dynamic, situation-specific reward function that adapts to the immediate requirements of the environment. For each categorized state, AHP is utilized to determine the most appropriate weight, aligning each state's success with its objectives within the reward structure. This results in a dynamic, situation-specific reward function that adapts to the immediate requirements of the environment. Our objectives include time efficiency, energy efficiency, obstacle avoidance, and human preference. The mathematical formulation is adopted from [23]. Moreover, the pairwise comparison matrix is weighted based on survey results and the references mentioned. The criteria and AHP weights, based on the states, are presented in Table I.

TABLE I THE AHP WEIGHTS REGARDING TO EACH STATE.

| Criteria/Objectives | $l_1$ | $l_2$ | $l_3$ | $l_4$ |
|---|---|---|---|---|
| Time Efficiency | 0.417 | 0.083 | 0.136 | 0.103 |
| Energy Efficiency | 0.417 | 0.083 | 0.191 | 0.137 |
| Obstacle Avoidance | 0.083 | 0.417 | 0.042 | 0.724 |
| Human Preference | 0.083 | 0.417 | 0.631 | 0.035 |

*3) Similarity-based experience replay*

For sample efficiency and avoiding non-stationary in the learning, we introduce a similarity-based experience replay mechanism. This approach aims to balance the learning from contextually relevant experiences with the diversity offered by random experiences, thereby mitigating the risk of overfitting while optimizing the learning efficacy.

Each experience in the replay buffer is represented as a tuple $E = (s, a, r, s', l)$, where s denotes the current state, the action taken, r the received reward, $s'$ the subsequent state, and $l$ the categorical label indicating the experience's contextual scenario. We categorize experiences based on operational states. To enrich the learning process with both relevant and diverse experiences, our mechanism concurrently retrieves two types of experiences for comparison: one that is like the current operational context ($E_s$) and another selected randomly from the buffer ($E_r$), irrespective of its contextual label. The Temporal Difference (TD) error for each experience, defined as

$$\delta(E) = r + \gamma max a' Q(s', a') - Q(s, a) \quad (3)$$

serves as the criterion for selecting which experience to prioritize in the learning update. Here, $Q(s, a)$ represents the predicted Q-value for action $a$ in state $s$, and $\gamma$ is the discount factor. During each learning iteration, we compute the TD errors for both Es and Er. The action associated with the experience yielding the higher absolute TD error is chosen for the update process, leveraging the premise that higher TD errors indicate a greater potential for learning. This dual-experience comparison ensures that the UAV's decision-making model benefits from a balanced exposure to both contextually pertinent and diverse experiences, fostering a robust learning framework that is adaptable to the dynamic and complex nature of SAR operations.

## IV. EXPERIMENTS AND RESULTS

*A. Experiment environment*

The experiment environment is constructed based on MATLAB 2023b to implement UAV autonomous navigation in 3D complex environments considering human centric factors. The simulation covers a square area of 200x200 m, with obstacles represented as ten cylinders of random radius and a fixed height of 50 m, simulating the maximum altitude for UAV operation. The size and position of each obstacle, along with the survivor's location and the UAV's starting point, are randomized at the onset of each episode to ensure varied and challenging training scenarios. Visual cues within the simulation are color-coded, with red indicating the UAV's initial position and blue denoting the survivor's location (see Fig. 8). This setup is aimed at replicating varied SAR scenarios, incorporating a total of 5000 episodes for the training process, each with a maximum step size of 300.

Central to our simulation is the modeling of human survivors, influencing their reactions to UAVs' varying speeds and altitudes. Drawing from seminal research [8, 9], we incorporate detailed reaction models that simulate real human responses to UAVs in SAR missions. These models account for the speed variation of the UAV and the resulting drift distance of humans when approached by UAVs at different speeds. To accurately reflect the zones of human comfort and discomfort, our simulation establishes zones based on the safety perceptions and drift distances outlined in the referenced literature. The UAV incurs penalties for entering discomfort zones or failing to maintain an optimal proximity. The simulation also accounts for mission failure conditions related to survivor position drift exceeding 20 m or approaching speeds of over 20 m/s within a 20 m proximity of the human.

While the UAV approaches a human at a speed of 15 m/s within a proximity of 20 m, we simulate a dynamic model that gauges human reactions based on the UAV's closeness. The behavior of this model is governed by the speed of the UAV, wherein the human's velocity adjusts proportionally to the UAV's velocity, ensuring that the model accurately reflects realistic human responses to the UAV's movements. The performance of proposed algorithm is compared with two benchmark algorithms DDPG and TD3 [8]. It should be noted that DDPG and TD3 are among the most used machine learning algorithms for autonomous decision making in continuous environments. The details of hyperparameters are listed in Table II.

TABLE II THE HYPERPARAMETERS OF THE PROPOSED ALGPRITHM.

| Actor/ Critic Learning Rate | 1e-4/1e-3 | Noise Clip | 0.5 |
|---|---|---|---|
| $\gamma$ | 0.99 | Exploration Noise | 0.1 |
| Replay Buffer Size | 500,000 | Batch Size | 128 |
| Policy Update Frequency | 2 | Policy Noise | 0.2 |
| Output Activation Function | Tanh | Activation Function | ReLU |
| Actor/ Critic Hidden Layers | [400,300] | $\tau$ | 0.005 |

### A. Training Results

In the training environment, there are 12 cylindrical obstacles with a random size and fixed altitude. Fig. 7 shows the training data's success ratio and average cumulative reward function of the proposed method compared with TD3 and DDPG. During the training process, we train the model with different random seeds for 5000 episodes and record the average rewards obtained for each episode.

As illustrated in the figures, proposed algorithm outperforms others by achieving the highest average reward and success rate. During the training phase, it demonstrated a significantly faster convergence rate compared to benchmark algorithms. From the outset, our algorithm exhibited superior cumulative rewards, a testament to the effectiveness of integrating the AHP and label-based similarity experience replay. This integration not only enhances sample efficiency by concentrating on the most relevant experiences but also dynamically adjusts reward structures in response to the operational context, facilitating quicker convergence towards optimal policy stabilization. This approach significantly reduces the number of episodes required to achieve policy stabilization compared to both DDPG and TD3, highlighting the efficiency of our customized method in speeding up the learning process. The graph indicates that the proposed method reduced variance over episodes, suggesting that the method not only achieves better performance in balancing multi-objective optimization tasks but also demonstrates improved learning stability and robustness.

This performance is underpinned by the integration of the AHP and label-based similarity experience replay. Such integration not only boosts sample efficiency by focusing on the most pertinent experiences but also allows for the dynamic adjustment of reward structures based on the current operational context, leading to quicker optimal policy stabilization. This method requires fewer episodes to achieve policy stabilization compared to DDPG and TD3, showcasing our approach's effectiveness in expediting the learning process.

As illustrated in Fig. 7, a key factor in our method's enhanced success rate compared to DDPG and TD3 is its ability to adaptively prioritize mission-critical objectives, particularly in high-risk scenarios. This adaptability, combined with the strategic application of similarity-based experience replay, empowers the UAV with a heightened capacity for informed decision-making. By drawing on contextually relevant past experiences, our algorithm not only navigates safer paths but also accelerates learning from similar scenarios, significantly improving success rates.

### A. Test results

In our study, we tested our proposed algorithm in a simulated environment populated with fifteen cylindrical obstacles of varying radius, placed randomly in each episode. This setup was designed to challenge the algorithm's capacity to generalize across different scenarios: E1, where human-centric factors were omitted; E2, incorporating human-centric considerations; and E3, positioning humans adjacent to obstacles. We benchmarked our algorithm against the DDPG [14] and TD3 algorithms, evaluating each on 3 key metrics: average success rate, collision rate, and average reward.

Table III presents the results, where our algorithm notably excels, achieving success rate, significantly outperforming DDPG's and TD3's.

TABLE III PERFORMANCE COMPARISON FOR DIFFERENT ENVIRONMENTS.

| Method | DDPG | | | TD3 | | | Proposed method | | |
|---|---|---|---|---|---|---|---|---|---|
| Metric | E1 | E2 | E3 | E1 | E2 | E3 | E1 | E2 | E3 |
| SR (%) | 79.3 | 68.2 | 54.6 | 88.4 | 82.5 | 76.9 | 92.7 | 89.1 | 85.3 |
| CR (%) | 14.7 | 22.6 | 33.5 | 7.3 | 11.5 | 15.8 | 3.9 | 6.3 | 8.1 |
| AR | 2.96 | 2.64 | 2.01 | 3.56 | 3.36 | 2.82 | 3.98 | 3.69 | 3.42 |

This superior performance is particularly pronounced in the most challenging scenario, E3, demonstrating our algorithm's adept handling of complex human-centric factors. This is achieved by strategically prioritizing rewards based on state categories, which enhances decision-making in scenarios with increased complexity. Unlike the benchmarks, which see reduced success and increased collision rates in E3, our method maintains a high success rate and demonstrates exceptional navigational safety, with the lowest collision rates among the competitors. This success in minimizing collisions, crucial for SAR operations, is due to our method's effective use of the AHP and similarity-based experience replay, enabling it to navigate complex terrains safely and adapt to the diverse demands of SAR missions.

Furthermore, our algorithm's minimal decline in success rate in E3, despite the increased operational complexity, underscores its robustness and effectiveness. The improvement in average reward signifies not only enhanced mission performance but also the algorithm's capability to balance and achieve a wide range of objectives, highlighting its potential to transform UAV strategies in SAR contexts.

While our method improves path length compared to DDPG, it slightly trails behind TD3 in this aspect. This reflects a strategic focus on various priorities, such as safety and adaptability over sheer speed, especially in scenarios involving proximity to survivors and obstacles, where immediate time efficiency is secondary. The Fig. 8 illustrates the path generated by the proposed algorithm for the E2.

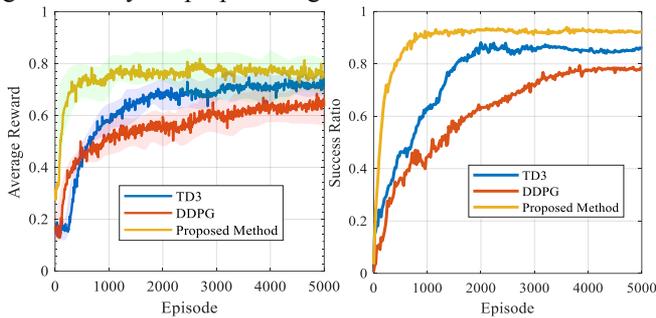

Figure 7. Normalized average reward function and success ratio for training environment.

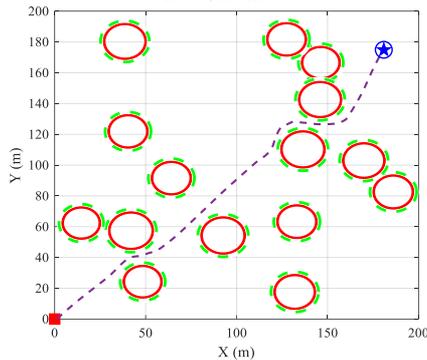

Figure 8. The generated path by the proposed method for case E2.

## V. CONCLUSION AND FUTURE WORKS

This study introduces a novel integration of the AHP and similarity-based experience replay with the TD3 framework to enhance the efficiency and effectiveness of drone-assisted SAR operations. The proposed methodology demonstrates superior performance in operational efficiency and learning efficacy compared to TD3 and DDPG, particularly in SAR missions. This research contributes to the field of UAV-assisted SAR operations by offering a comprehensive, human-centered approach that addresses both the technical and practical challenges of using drones in disaster relief and emergency response efforts.